# A Common Operating Picture Framework Leveraging Data Fusion and Deep Learning

Benjamin Ortiz, David Lindenbaum, Joseph Nassar, Brendan Lammers, John Wahl, Robert Mangum, Margaret Smith, and Marc Bosch

*Accenture Federal Services, Arlington, VA, USA.*

**Abstract**: Organizations are starting to realize of the combined power of data and data-driven algorithmic models to gain insights, situational awareness, and advance their mission. A common challenge to gaining insights is connecting inherently different datasets. These datasets (e.g. geocoded features, video streams, raw text, social network data, etc.) per separate they provide very narrow answers; however collectively they can provide new capabilities. In this work, we present a data fusion framework for accelerating solutions for Processing, Exploitation, and Dissemination (PED). Our platform is a collection of services that extract information from several data sources (per separate) by leveraging deep learning and other means of processing. This information is fused by a set of analytical engines that perform data correlations, searches, and other modeling operations to combine information from the disparate data sources. As a result, events of interest are detected, geolocated, logged, and presented into a common operating picture. This common operating picture allows the user to visualize in real time all the data sources, per separate and their collective cooperation. In addition, forensic activities have been implemented and made available through the framework. Users can review archived results and compare them to the most recent snapshot of the operational environment. In our first iteration we have focused on visual data (FMV, WAMI, CCTV/PTZ-Cameras, open source video, etc.) and AIS data streams (satellite and terrestrial sources). As a proof-of-concept, in our experiments we show how FMV detections can be combined with vessel tracking signals from AIS sources to confirm identity, tip-and-cue aerial reconnaissance, and monitor vessel activity in an area.

# I. Introduction

Overhead imagery systems capable of automatic target recognition (ATR) and/or visual sense making are not enough for gaining full situational awareness. Thanks to the proliferation of different types of sensor, other data modalities are available for intelligence collection. Combination of different data modalities for collective gain is an area underexplored within the ATR community.

With this in mind, we have developed a video analytics service platform (VASP) to accelerate data fusion for improved Artificial Intelligence (AI) / Machine Learning (ML) solutions for Processing, Exploitation, and Dissemination (PED). VASP is an extensible, algorithm agnostic, modular and layered platform as a service (PaaS) solution that enables data fusion and modeling at scale. VASP provides the toolset to rapidly build and deploy AI capabilities with the addition of multiple powerful analytics and visualization tools to help analysts increase operational insight and timely decisions from computer vision models. Our goal is to field a rapid toolset that moves beyond basic detection with reasoning functions that contextualize, correlate, and infer greater situational awareness through data fusion. Our data fusion framework is based on cross-processing of signal data. We use correlations between signals to fuse data and to establish source correspondence.

We have implemented an end-to-end functionality for the entirety of the video analytics and data fusion pipeline. VASP can ingest multiple data sources of varying format and complexity through flexible integration and ingestion points, including API connectivity, to create a single database for analysis. The Video and Sensor Analytics layer applies algorithms to the ingested video data (FMV, WAMI, CCTV / Fixed Camera, Open Source Video, etc.) to identify key objects, events, and sequences of interest. We have developed a library of algorithmic techniques for both fixed camera and drone collection.

In addition to video, VASP ingests streaming sources of non-video geo-rectified data (e.g. AIS, ADS-B, EW, SIGINT). The system stores the data for analysis and forensic review while, in parallel, creating correlations based on proximity to other events, both video and non-video. The architecture enriches the data through access to OSINT and web-scrapping to support both geo ("dots on a map") views and graph / social network analysis ("connections", "connections on a map") tools for greater insight. In this work, we have focused on processing and fusing video feeds, artifacts from its processing and Automatic Identification System (AIS) used to track maritime vessels.

# II. System Overview

Figure 1 shows the front-end common operating picture and some features implemented as part of VASP. The front-end provides the analyst the ability of gaining rapid awareness of the different sensors/signal sources being deployed in the field, (e.g. AIS and FMV data). The front-end is divided in functional areas: (1) data source - (e.g. AIS, FMV), (2) Detections filter, (3) Events log, (4) live cameras, (5) image similarity search, (6) Analytics, (7) common operating picture map with geolocated artifacts (e.g. events, detections).

*Data source* indicates all the different signal modalities that are being involved in the current scenario. *Detection filter* list the different objects of interest found and identified in the data sources as well as their amount in the current session. *Events log* tool allows to visualize and verify an event of interest found in FMV feeds. We have investigated and designed a system capable of detecting events and activities of interest that deviate from the baseline patterns of observation given FMV feeds in order to reduce the amount of redundant information produced by these systems as well as providing the human analyst a faster methodology to gain situational awareness. The exact frame where an event was detected along with a description of the event and geolocation information are available for a deeper examination. The platform also has a direct link to the live feeds for maintaining awareness. The platform is scalable and can support a different number of feeds as well as resolution and formats. Detections are also overlaid on top of the live feed for faster visual inspection and facilitate focus on certain activities. *Image similarity search* capability allows to inspect historical data to perform media retrieval and search of past events. The *analytics* section captures different activity in the system, from variations, changes and anomalies in the system activity, to object counts. Thus, enabling pattern-of-live analysis. Finally, the *common operating picture map* shows all activity (detections, events) geolocated using a mapping service API. Each visualization is an active link that allows further exploration of a particular object/event instance.

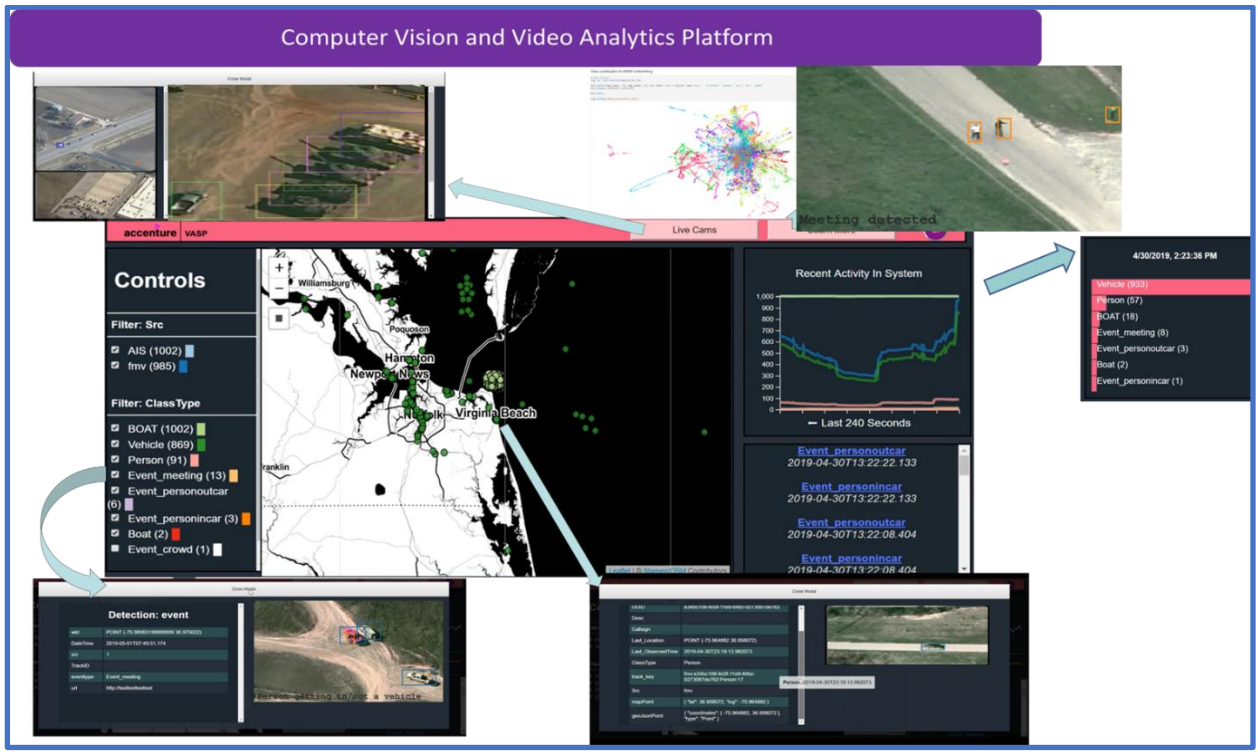

*Figure 1. Common Operating Picture Framework*

# III. AIS – FMV Fusion

Fusing video data sets and detections with all-source information for a clear picture of circumstances and dependencies. We use statistical methods to infer relationships between data sets and various objects / entities for geospatial, network, and timeseries analyses, with the potential for predictive modeling as the data baseline grows. Use cases may include leveraging HUMINT to verify person identity, SIGINT to layer selector detail and connect entities of interest, or something as simple as weather data to anticipate behaviors given environmental conditions.

Different data sources demand different approaches to data analysis. In our example, we primarily worked with two main data sources: one supplying the metadata provided by the Automatic Identification System (AIS) used to track marine vessels and one supplying Full Motion Video (FMV) from an Unmanned Aerial Vehicle (UAV).

The International Maritime Organization requires AIS devices be fitted aboard international voyaging ships with at least 300 gross tonnage and all passage ships regardless of size. We acquired data from the AIS devices from a live satellite feed provided by a partner organization. Data provided by AIS devices include the ship's Maritime Mobile Service Identity (MMSI), time, longitude, latitude, Course Over Ground, Speed Over Ground, and Heading. With this information we know the historical locations of a vessel as well as predict its future locations. This information is useful if an analyst is interested in a particular vessel.

However, if an analyst is interested in a large number of vessels, then it quickly becomes cumbersome to identify ships of interest. We can supplement the provided AIS data with event detection. Event detection can take on many forms, but we focus on three cases: disappearance, appearance, off-course, and co-location. Because AIS data is required for certain ships, if the AIS transmission disappears for a prolonged period, then this can possibly mean a region with weak GPS signals, a vessel with faulty equipment, or even a deliberate drop of signal. If the analyst is interested in a particular region, then geofencing can be used to detect when ships are entering, exiting, or projected to do either for a given bounding box. Also, because we can reasonably predict where a ship is heading, if a ship's current location is not on the same course as was predicted, then we may need to investigate a reason for such a drastic change. If two ships are located within a certain distance of each other in the water, then this could signify a crash or a physical interaction between vessels.

Unlike AIS data, FMV data is unstructured in nature. It traditionally requires the use of human analysts to manually summarize and annotate relevant objects or activities that occur in the time frame of the video. However, advances in Computer Vision (CV) machine learning capabilities allow for the use of historically annotated FMV data to be used to confidently detect objects of interest without direct human supervision. Objects detected with CV can be verified by a human observer by a labelled bounding box provided by the CV algorithm. Once the objects are detected, they can be further utilized to detect activities [1]. For instance, if the object of interest is a "person," then a heuristic can be created to identify the overlap of two "person" bounding boxes to signify a "meeting" or the overlap of three or more "person" bounding boxes to signify a "gathering".

With use of CV techniques, we can label FMV data and therefore add structure to unstructured data. This data is enhanced by the fact that it comes from a UAV. UAVs contain metadata including latitude, longitude, altitude, and timestamped information. The coordinate and time data of the UAV can be combined with the newly structured data of the FMV to accurately locate an object or an event.

While our two disparate data sources are different in nature – one provides structured data while the other provides unstructured data – we can derive similar information from them, primarily geolocated event data. The goal is to unify the two datasets in a single data source. This can be done with the use of a microservice architecture. Our microservice architecture processes the two data sources independently, then consolidates the two distinct data layers into a single data layer, then makes the consolidated accessible via a RESTful API (see figure 2 for more details).

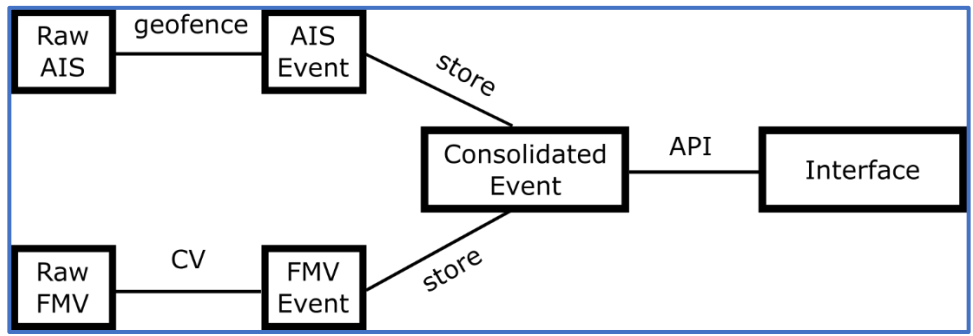

Figure 2. AIS and FMV fusion process.

Through a series of correlations, we can associate FMV detections and AIS so that different operations can occur. Vessel AIS verification, through tip-and-cue of an FMV from AIS can be done after a new vessel has entered a geofence area. Figure 3. Once visually verified by the FMV and the CV engine, the vessel can be moved to the list of validated vessels. This can help in the identification of AIS tampering, fraud, by visually and through computer vision to validate that the vessel matches its AIS signature.

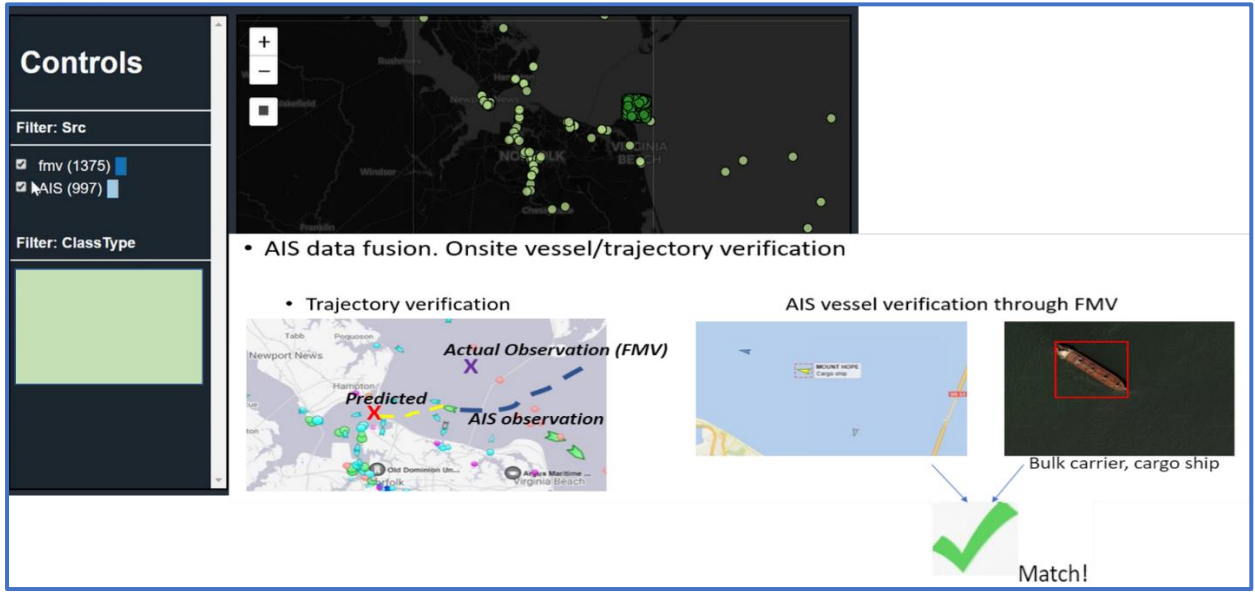

*Figure 3. Example of AIS - FMV fusion*

# IV. Image Similarity Search

A critical feature in any video analytics tools is the ability to search through historical data for "similar" events. Due to the large size and volume of data that an FMV system produces, it becomes very difficult for an analyst to recall the exact instance where a similar event was captured. Instead, we propose to automate this and fold it into a capability to allow analysts and users to quickly find similar events. We approached this by designing a system that performs image-to-image search based on mathematical distance between image features. Essentially providing analysts the ability to search an image of an object, or a specific scene, against a reference image database based on the similarity of the photo or video snapshot. This enables search without having to tag or annotate the historic image data. Search can help analysts determine if an object has been previously detected, is an anomaly, and quantifies / clusters similar objects to manage navigation of large-scale data sets. This is a highly unconstrained and unsupervised task. We have developed a system that extracts features of the query image and quickly compares and returns the list of closest matching entries in the database. Because user searches are semantical rather than visual, we needed to develop a model that captured the semantics of the query image. For this reason, we used a deep neural network that extracts features (VGG-19 [2]) that were learned through the process of object detection. In other words, we leveraged the backbones used to detect objects from the FMV to extract features used to perform the image search. These features are then passed through the Uniform Manifold Approximation Projection (UMAP) [3] algorithm that reduces the dimensionality of the data while creating a representation topology of the database that makes search operations more efficient. UMAP projects the data into a well-structured representation space that preserves semantics so that similarity is referred to semantics rather than visual. See figure 4 for an example projection. For example, if the query image is a boat surrounded by water, the expected output from a search operation would be same or similar boats, regardless of the scale and surrounded by similar visual content, not simply copies of a boat at the same scale. Or just visual objects that have similar color composition to the query example. Like it is shown in figure 5 as one of our results.

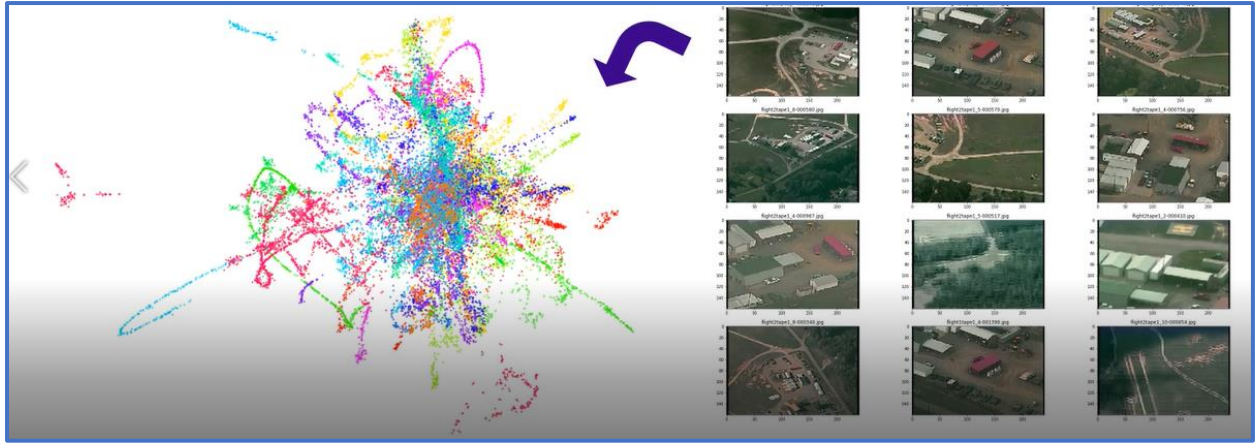

*Figure 4. UMAP mapping for image search.*

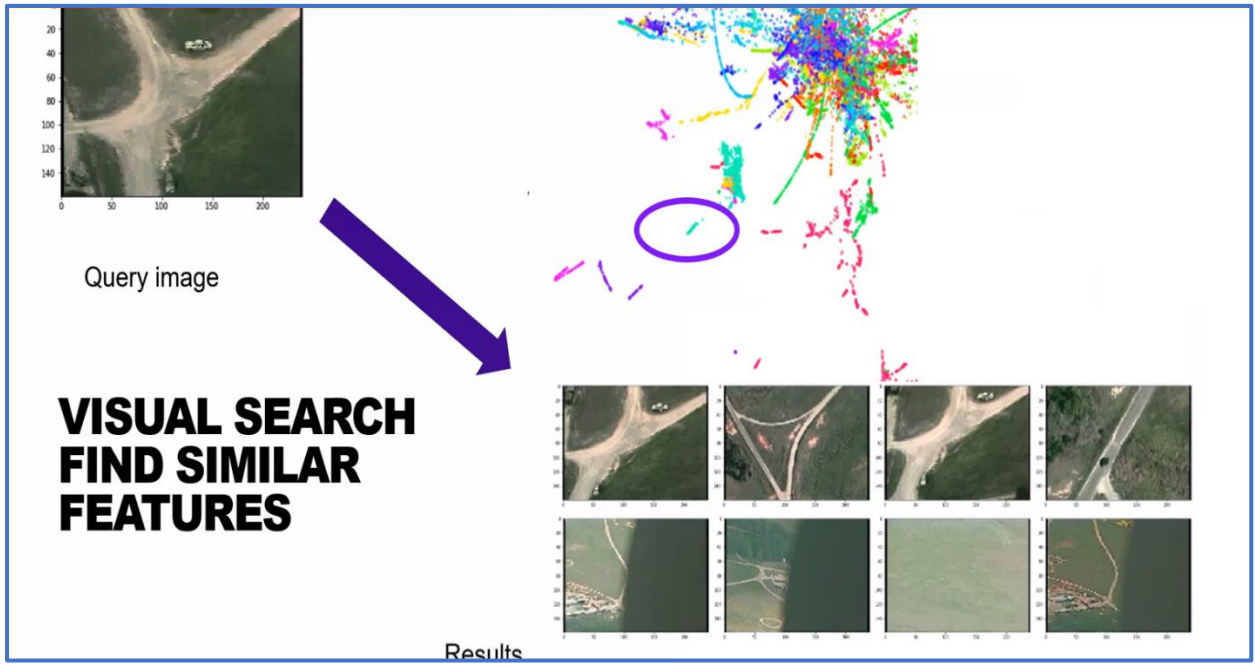

*Figure 5. Similarity search result.*

# IV. Other Features

As mentioned earlier, besides data fusion and similarity search, VASP provides solutions for Processing, Exploitation, and Dissemination (PED). We have investigated and designed a system capable of detecting events and activities of interest that deviate from the baseline patterns of observation given FMV feeds in order to reduce the amount of redundant information produced by these systems as well as providing the human analyst a faster methodology to gain situational awareness. A series of analytics monitor the scene by keeping track of object counts, and object interactions. If these object interactions are not declared to be commonly observed in the current scene, the system will report, geolocate, and log the event. Examples of events of interest include identifying a gathering of people as a meeting and/or a crowd, alerting when there are boats on a beach unloading cargo, increased count of people entering a building, people getting in and/or out of vehicles of interest, alerting when a vehicle goes from static to moving or changes speed, etc. Figure 6 shows examples of events detected and presented to the analyst for inspection. Similarly, in figure 7 the live feed monitoring user interface can be seen, where the user is provided with real-time access to the processed feed for a quick inspection highlighting certain content.

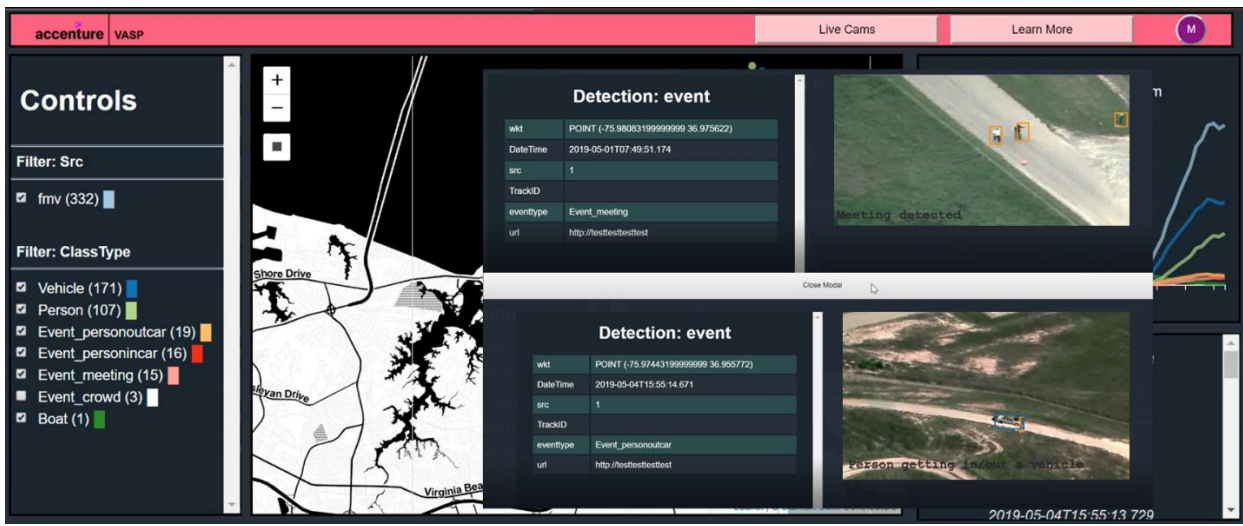

*Figure 6. Event detection through the common operating picture.*

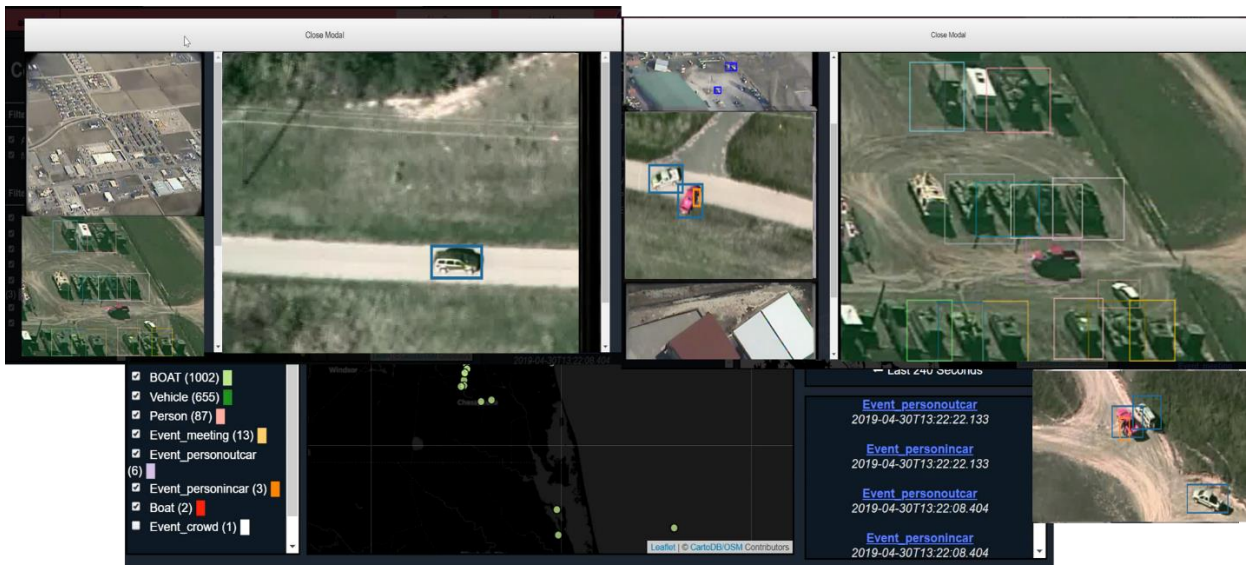

*Figure 7. Live real-time processed FMV feed access and monitoring.*

# V. Conclusion

In this paper, we have provided an overview of our common operating picture. We have described a data fusion framework for accelerating solutions for Processing, Exploitation, and Dissemination (PED). We built a platform as a collection of services that extract information from several data sources (per separate) by leveraging deep learning and other means of processing for data fusion, image search and FMV processing that include events of interest. These events are detected, geolocated, logged, and presented into a common operating picture. This common operating picture allows the user to visualize in real time all the data sources, per separate and their collective cooperation. Our initial data fusion process involved AIS and FMV data, we showed how this capability can be used for tip-and-cue UAV systems for vessel verification, or for verifying vessel trajectory.